# CHAMP: CROWDSOURCED, HISTORY-BASED ADVISORY OF MAPPED PEDESTRIANS FOR SAFER DRIVER ASSISTANCE SYSTEMS

**Ross Greer**
**Lulua Rakla**
**Samveed Desai**
**Afnan Alofi**
**Akshay Gopalkrishnan**
**Mohan Trivedi,** *Faculty Mentor*
University of California San Diego
USA

Student Safety Technology Design Competition: Advanced Driver Assistance Systems

## ABSTRACT

Vehicles are constantly approaching and sharing the road with pedestrians, and as a result it is critical for vehicles to prevent any collisions with pedestrians. Current methods for pedestrian collision prevention focus on integrating visual pedestrian detectors with Automatic Emergency Braking (AEB) systems which can trigger warnings and apply brakes as a pedestrian enters a vehicle's path. Unfortunately, pedestrian-detection-based systems can be hindered in certain situations such as nighttime or when pedestrians are occluded. Our system, CHAMP (Crowdsourced, History-based Advisories of Mapped Pedestrians), addresses such issues using an online, map-based pedestrian detection system where pedestrian locations are aggregated into a dataset after repeated passes of locations. Using this dataset, we are able to learn pedestrian zones and generate advisory notices when a vehicle is approaching a pedestrian despite challenges like dark lighting or pedestrian occlusion. We collected and carefully annotated pedestrian data in La Jolla, CA to construct training and test sets of pedestrian locations. Moreover, we use the number of correct advisories, false advisories, and missed advisories to define precision and recall performance metrics to evaluate CHAMP. This approach can be tuned such that we achieve a maximum of 100% precision and 75% recall on the experimental dataset, with performance enhancement options through further data collection.

## INTRODUCTION

According to the latest data from the National Highway Traffic Safety Administration, 6,516 pedestrians died and 54,769 were injured in traffic crashes in the United States in 2020 [1]. From 2000 to 2020, there has been an increase by 42% in the number of pedestrian fatalities on public roadways despite all the development in Vehicle and road safety. The majority of pedestrian traffic deaths occurred in urban areas (80%), on open roads (75%) rather than at intersections (25%). The data showed that the primary factor of pedestrian traffic deaths was the failure to yield right of way, with 50% [1]. Due to the prioritization of infrastructure designed for the convenience of cars over the last few decades, safe and convenient pedestrian infrastructure has been dramatically reduced. As a result, outside of urban areas with high walk-scores, pedestrian activity is sparsely distributed. However, in most cases, pedestrian patterns emerge that may or may not align with existing pedestrian paths or marked crossings. In these areas, it can be difficult for new drivers, or drivers unfamiliar with certain neighborhoods to understand where to expect pedestrians.

For pedestrian detectors, vehicles often use cameras or radar to check if there are pedestrians ahead of them. If a pedestrian is in the car's path, then a warning will be triggered by the vehicle to alert the driver. These warnings can vary from something like a popup message on a dashboard, alert sound, or a seat vibration, if possible. These

**Acknowledgements:** The authors would like to thank NHTSA and ESV for the opportunity and feedback in developing CHAMP.

pedestrian detectors and alert systems can be combined with Automatic Emergency Braking (AEB) to avoid colliding with a pedestrian.

Automatic Emergency Braking (AEB) systems have improved pedestrian safety by reducing vehicle velocity to avoid or mitigate effects of collision, an ongoing risk when pedestrians are present at both marked and unmarked crossings [2, 3, 4, 5] and intersections. Haus et al. [2] show that all AEB models in their study were actually more effective when combined with driver braking, as opposed to autonomous behavior alone. Despite their effectiveness in certain situations, AEB systems are less effective in common driving scenarios. Cicchino [3] and a recent AAA study [4] show evidence that AEB systems are less effective in dark conditions, when vehicles are turning, and when pedestrians appear suddenly from occluded locations - all very common scenarios for pedestrian encounters. In such situations, vision-based pedestrian detectors are not as effective because these situations lack visual cues for the vehicle to recognize an approaching pedestrian.

To address this performance gap and implement these motivating principles, we propose a system that advises the driver when they are in the likely presence of pedestrians. We name this system CHAMP, which stands for Crowdsourced, History-Based Advisories of Mapped Pedestrians. Rather than being based on single-instance detections, the proposed system is an online, map-based approach, where pedestrian behaviors are aggregated and learned from repeated passes of single-camera vehicles. This repetition mitigates the effects of detection failure cases and environmental changes (lighting, occlusion, vehicle path). From the generated map, statistical pedestrian crossing patterns are inferred. Finally, an advisory threshold is created which can be tailored to driver preference or safety standard, with a possible range from activation if a pedestrian has ever crossed in the area, to activation only in areas where pedestrian activity matches or exceeds the flow of a high-traffic signalized crosswalk.

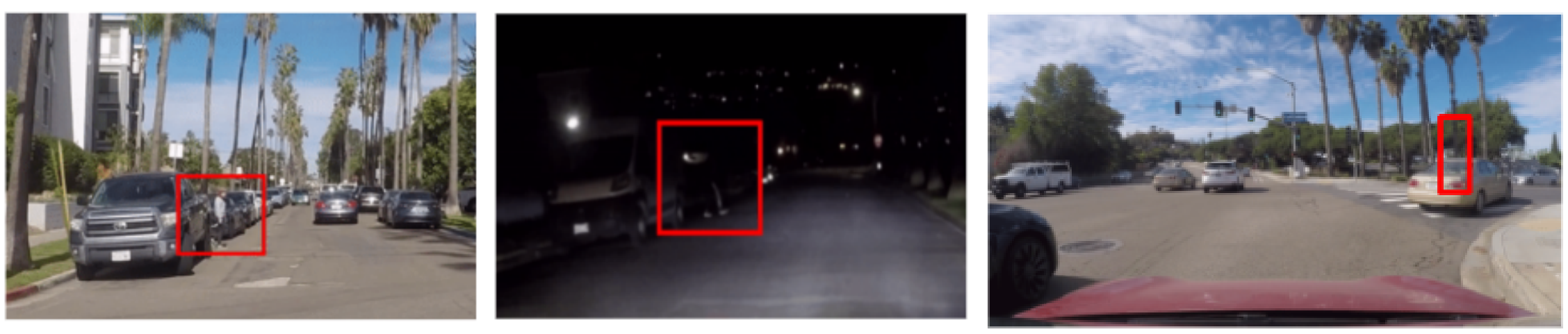

***Figure 1. CHAMP aims to tackle safety critical scenarios like occluded pedestrians (top left) , dark lighting conditions (top right) and blind right turns (bottom)***

Such an advisory system allows for improved driver vigilance in safety-critical pedestrian situations, improving opportunities for braking where AEB systems may fail or benefit from driver cooperation. In particular, this map-based approach will help vehicle safety in conditions such as driving at night time or generated warnings even when a pedestrian is occluded.

**RELATED RESEARCH**

Current studies that explore pedestrian detection can be categorized in several ways. First, some studies focus solely on creating pedestrian detection datasets, whether the focus be on a large dataset that reflects a variety of conditions involving pedestrians [6] or a focus solely on pedestrian detection in monocular images in urban environments [7]; however, a weakness of these datasets is that not all of them are entirely human-annotated. Second, other studies can be categorized by the conditions in which they attempt to detect and classify pedestrians, such as studies that use deep learning techniques and CNNs to detect pedestrians specifically in hazy weather conditions [8] and a variety of weather conditions [9], studies that use SVM for detection and Kalman filters for tracking of pedestrians in low-visibility conditions at night [10], and studies that use motion estimation and texture analysis techniques in order to detect pedestrians in crowded urban environments specifically [11]; however, many of these studies solely focus on detection of existing pedestrians, rather than predicting the likely movement trajectories of pedestrians.

Other studies that focus on predicting the movement of pedestrians can be categorized based on the techniques that they employ; Asahara *et al.* [12] use a mixed Markov-chain model to predict pedestrian trajectories, Particke *et al.* [13] use an advanced Kalman filter called a Multi-Hypotheses filter to model pedestrian movements, Song *et al.* [14] use a technique involving tensors to represent features of pedestrians in crowded environments and a convolutional LSTM to predict pedestrians' trajectories, and Keller *et al.* [15] use stereo-vision based path prediction for pedestrians. However, some of these studies only have a limited number of pedestrian movements that they consider; for instance, Keller *et al.* [15] only consider pedestrian movement lateral to the vehicle where the pedestrian stops and when the pedestrian continues to walk past the vehicle. Additionally, many of these studies are based on a single-instance detection of pedestrians; by contrast, our system CHAMP employs a method of aggregating pedestrian behaviors after repeated detections and passes.

## METHODS

### Fleet or Repeat

Our work addresses these issues through two principles, which we nickname "Fleet" and "Repeat", meant to show that long-term patterns in scene agent behavior can be uncovered using repeated samplings of a geographic location, either in a spatial sense (e.g. covered by a fleet of vehicles, "Fleet") or temporal sense (e.g. repeated visits from a single car, "Repeat").

As two examples of this method of repeated sampling to infer real-world patterns, Morris and Trivedi [16] show that repeated polls of detected vehicles on roadways and intersections over time yield clear lane patterns and turning flows, without any sampling of the actual infrastructure. Greer and Trivedi [17] show that repeated polls of a pedestrian intersection allow to map and learn clear crosswalks without sampling the infrastructure itself. Such long term-data will allow us to learn behavioral patterns of pedestrians and how they interact with road infrastructures, and both methods make use only of detected dynamic scene agents (vehicles and people) to infer behavioral patterns. Further, repeated visits also provide a secondary view of an area that may have been a blind spot (either in direction or in time), as illustrated in Figure 2.

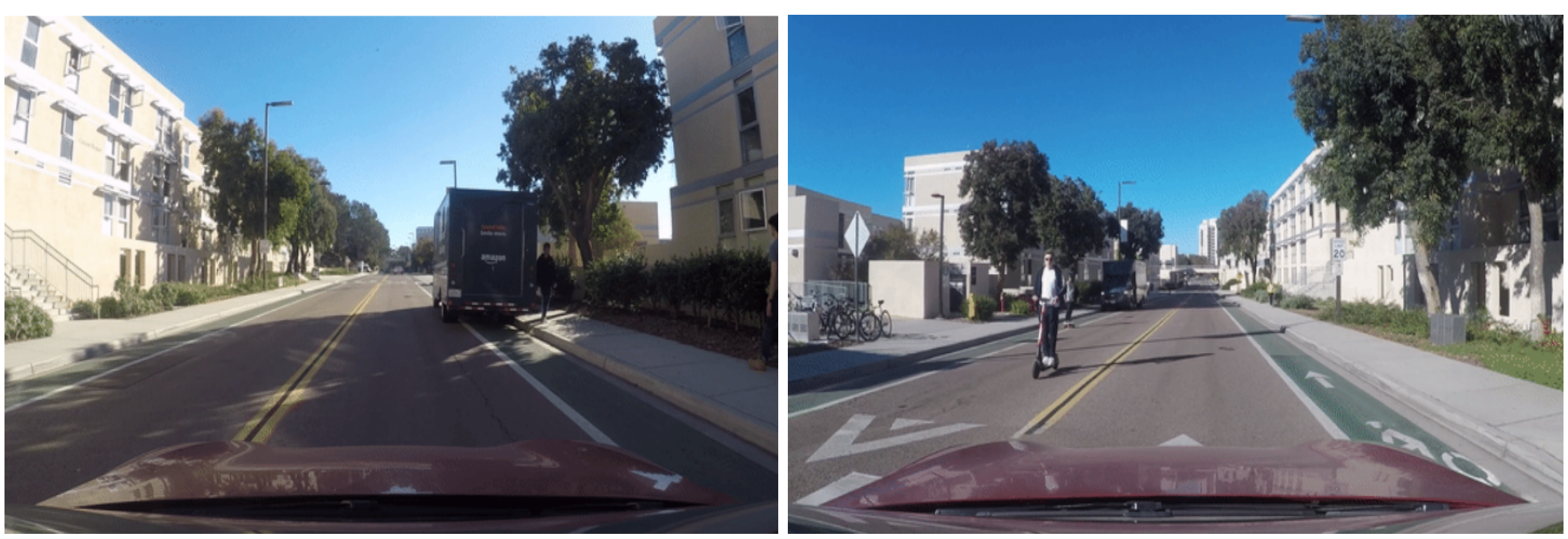

***Figure 2. A large truck occludes a view of pedestrians from either side of the truck, but having a vehicle positioned on both ends of the truck would capture enough information to remove this blind spot in a spatial sense. Similarly, returning to capture a new image after the truck has left the area would allow for resolution of the blind spot in a temporal sense. Such temporal blind spots may also be induced by poor lighting. The need to resolve these blind spots motivates a method of "Fleet or Repeat", whereby repeatedly sampling scenes we can detect pedestrians occluded from specific single views.***

## System Design

The CHAMP system consists of 3 main parts as illustrated in Figure 3:

1. Pedestrian Location Association
2. Nearest Pedestrian Hotspots Search
3. Advisory Issuance

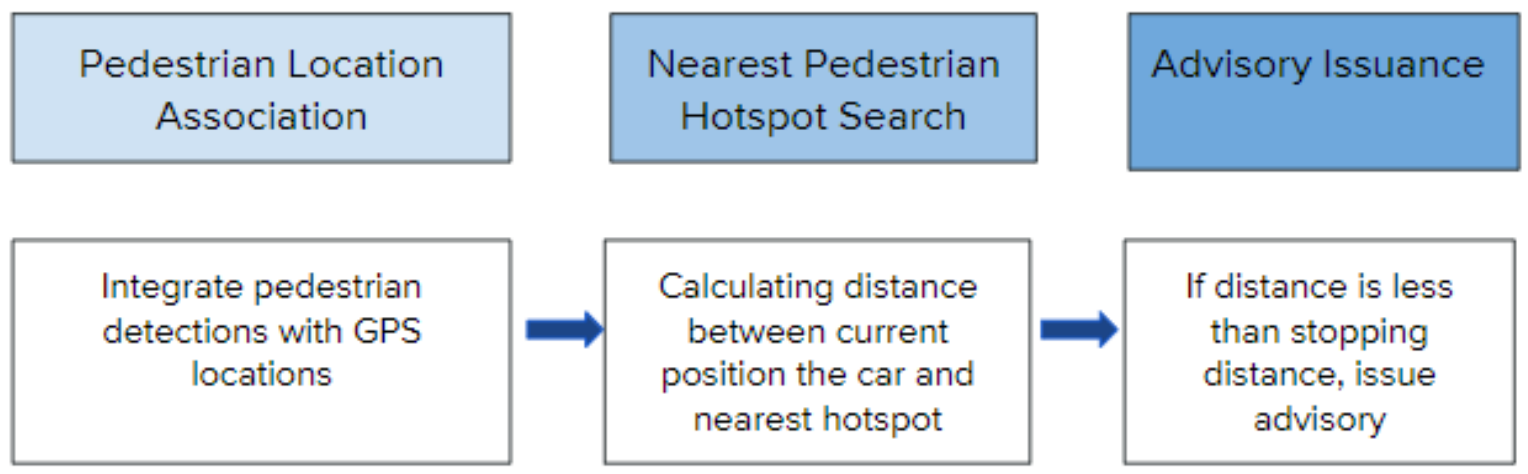

***Figure 3. System design of CHAMP, with three stages: location association, nearest hotspot search, and advisory issuance.***

**Pedestrian Location Association** CHAMP takes in the GPS information received from the drive training data and splits it into intervals of 1 second each, giving *p* location coordinates per second (dependent on the GPS recording rate). CHAMP then associates the $k^{th}$ position interval with all the pedestrians which were detected in the $k^{th}$ time frame, for that clip. Our system then takes the median of the *p* location coordinates of the vehicle associated with the $k^{th}$ position interval and stores the pedestrian count for that median location. Finally, CHAMP creates a cluster over these repeated samples of vehicle locations, wherever the pedestrians were detected. This algorithm is repeated as further area is covered and data is collected, scaling to add new pedestrian nodes.

**Finding Nearest Pedestrian Hotspots** At inference time, our algorithm uses a ball-tree-based approach using haversine distance to efficiently calculate the distance between the car position and the nearest pedestrian location $d_{car,ped}$. This approach uses a binary tree with a hierarchical structure to effectively compute the nearest neighbor in a multidimensional space, thus reducing the overall latency during test time. Our algorithm also checks if the pedestrian is in the field of view of the car by calculating the heading angle $\theta_{heading}$ between the current position of the ego vehicle and the nearest pedestrian, so that it can ignore pedestrians behind the vehicle.

**Issuing Advisory** CHAMP then uses stopping distance [18], which incorporates velocity as a proxy to Time-To-Collision (TTC), to calculate the radius around the car to check for pedestrians. To allow for additional safety, we also use a multiplicative offset, *b*. Stopping distance *s* is defined as

$$s = b \cdot \frac{(0.278 * t * v) + v^2}{254 * (f + G)} \quad \text{(Equation 1)}$$

,

where *t* is reaction time, *v* is velocity of the car (km/h), *f* is the coefficient of friction, and *G* is the slope. For our calculations, we used the approximated values *f* = 0.7 (assuming a dry road) and *G* = 0 (assuming most of our roads are flat without any uphill/downhill).

To reduce repetitive computations, CHAMP performs all the above calculations every K meters (where K is the sampling distance). K is a hyperparameter which we can tune according to observed performance considering vehicle speeds, GPS sampling rates, and pedestrian frequency.

During inference, if the distance between the car and the nearest pedestrian ($d_{car,ped}$) is less than the stopping distance (*s*) and the heading angle ($\theta_{heading}$) is less than 90 degrees, CHAMP issues an advisory to the driver. These advisories request the driver to be vigilant and keep their eyes on the road, making the driver aware that pedestrians may be present.

## METRICS AND EVALUATION

We gathered real-world driving data in the San Diego, California area. Data was captured by a front-facing GoPro camera mounted to the LISA-T testbed [19]. We aggregated 10,000 clips of 10 seconds each, which at a frame rate of 30 fps resulted in 3 million frames. We restricted the data to the La Jolla region of San Diego to facilitate testing and training with adequate coverage. After collection, expert human annotators marked all pedestrians in the frames. In addition to these annotations, we recorded the position of the car using GPS. In addition to training data collection, we took an additional 3 drives to generate test clips for evaluation of CHAMP, returning to previously visited areas at different times and from different directions. These drives captured scenarios of blind turns, occluded areas, and dark lighting conditions. CHAMP was then used to process the test data, producing advisory alerts at different locations. These advisory periods were then compared against human-defined ground truth scenarios to demonstrate the system quantitatively and qualitatively as illustrated in the next section.

### Quantitative Summary

CHAMP is evaluated using standard precision and recall metrics. Precision is the ratio of the correct advisories to the total advisories given, used to quantify the prevalence of false advisories (Equation 2). Recall is the ratio of the correct advisories to the total advisories the system should have given, used to quantify missed advisories (Equation 3).

$$Precision = \frac{|Correct\ Advisories|}{|Correct\ Advisories| + |False\ Advisories|} \qquad \text{(Equation 2)}$$

$$Recall = \frac{|Correct\ Advisories|}{|Correct\ Advisories| + |Missed\ Advisories|} \qquad \text{(Equation 3)}$$

As CHAMP is a safety system, Recall is a prioritized metric, as it is critical to be vigilant in zones where pedestrian behavior is prevalent and avoid missed advisories. There is a natural tradeoff between Recall and Precision, and in cases where the advisories become too frequent, the system should be re-tuned toward higher precision.

Another hyperparameter that can be modified to influence system performance is Sampling Distance, or the distance traveled between consecutive assessments for advisory. This controls how often we compute the nearest pedestrian from the current location of the car. As the nearest pedestrian does not change in short distances of 1-2 meters, sampling at lower frequency allows for reduced computational cost.

Table 1 summarizes the performance of the system using these metrics. Clip 1 contains a blind right turn, Clip 2 contains pedestrians occluded by vehicles before crossing, and Clip 3 contains dark conditions (nighttime). These clips were selected to test the system when faced with typical failure cases for pedestrian detection or AEB systems.

*Table 1*
*CHAMP Evaluation: Precision and Recall by Clip and Sampling Distance*

| Test Clip ID | Duration (min) | Scenario | Sampling Distance (m) | Precision | Recall |
|---|---|---|---|---|---|
| 1 | 2 | Blind turns | 2 | 0.27 | 0.75 |
| 2 | 2 | Occluded Pedestrians | 2 | 0.75 | 0.75 |
| 2 | 2 | Occluded Pedestrians | 5 | 1 | 0.25 |
| 3 | 2 | Nighttime Lighting | 2 | 0.105 | 0.72 |

In our examples, we find that increasing sampling distances affects recall negatively, as certain pedestrians may be missed within a 5 meter span between computations. Hence, we suggest a sampling distance of 2 meters (though this can be tuned further depending on the driving environment). To increase precision and reduce false advisories, we propose future development in denoising and thresholding steps, including (1) increased sampling to allow for filtering by frequency of pedestrian sightings in a given area, (2) incorporation of trajectory prediction models so that pedestrian data can be filtered to include only instances where pedestrians actually cross the street, and (3) stratification of advisories by time of the day (e.g. a school zone may issue advisories around the beginning or end of the school day, or a street with busy nightlife may issue advisories during nighttime drives).

**Qualitative Examples**

Figure 4 illustrates a scene from Clip 2 in the test data, featuring pedestrians emerging from an occluded area. During training, CHAMP has seen pedestrians in this area that are within the radius of the advisory distance and in front of the car, so CHAMP issues a vigilance advisory notice to the driver. This allows for the driver to be alert for the occluded pedestrian crossing the street.

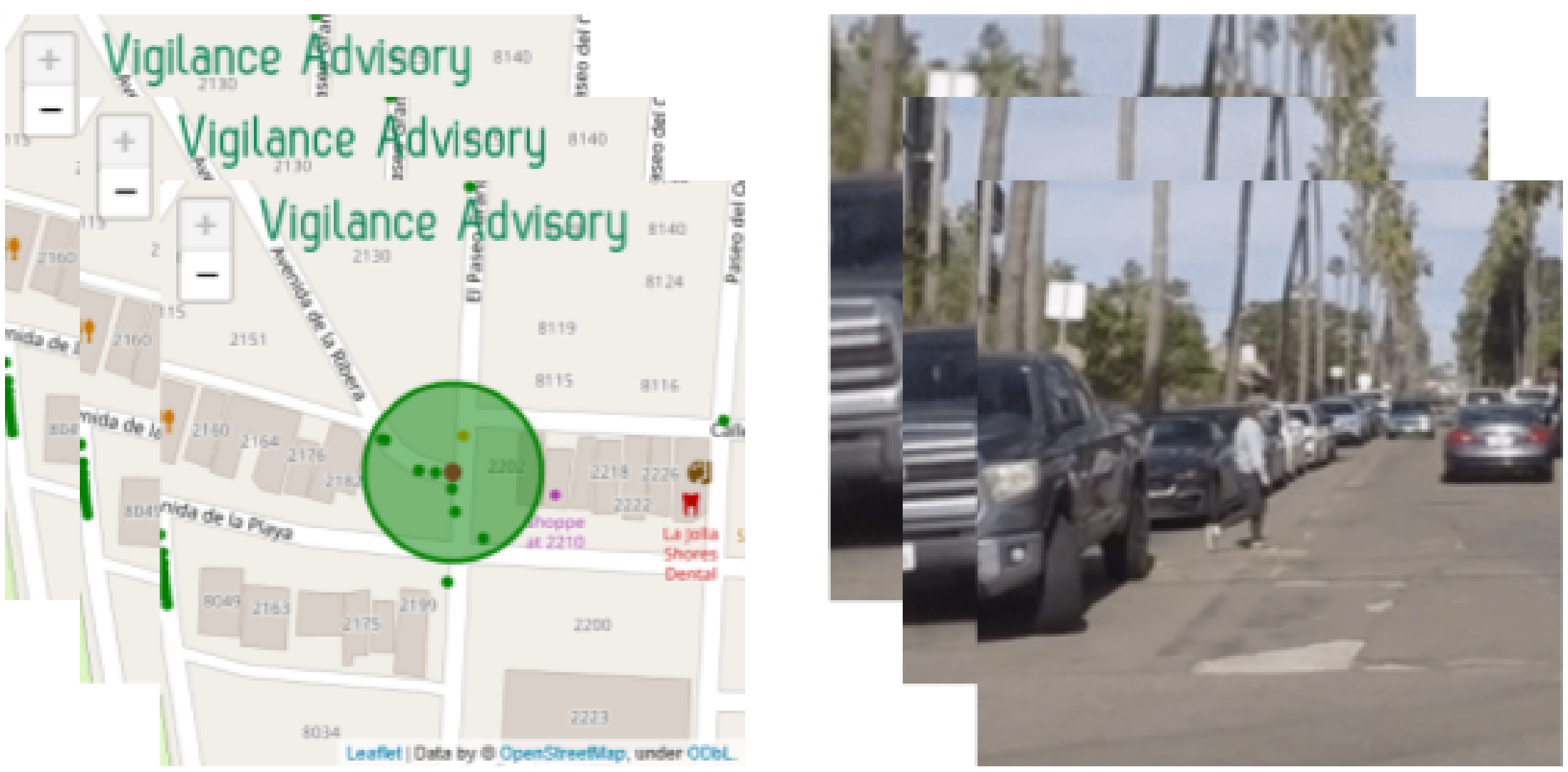

***Figure 3. CHAMP issues a warning in sequential frames to advise the driver that there may be pedestrians nearby, based on the training data. In the image on the left, the red dot represents the ego-vehicle, the green dots are pedestrians seen in the training set, and the yellow dot is the unseen pedestrian in the test data.***

Now that the driver has been advised by CHAMP, they can properly slow down and wait for the pedestrian to cross the street. In Figure 5, we can see that once we are behind any of the training points the vigilance advisory notice will turn off.

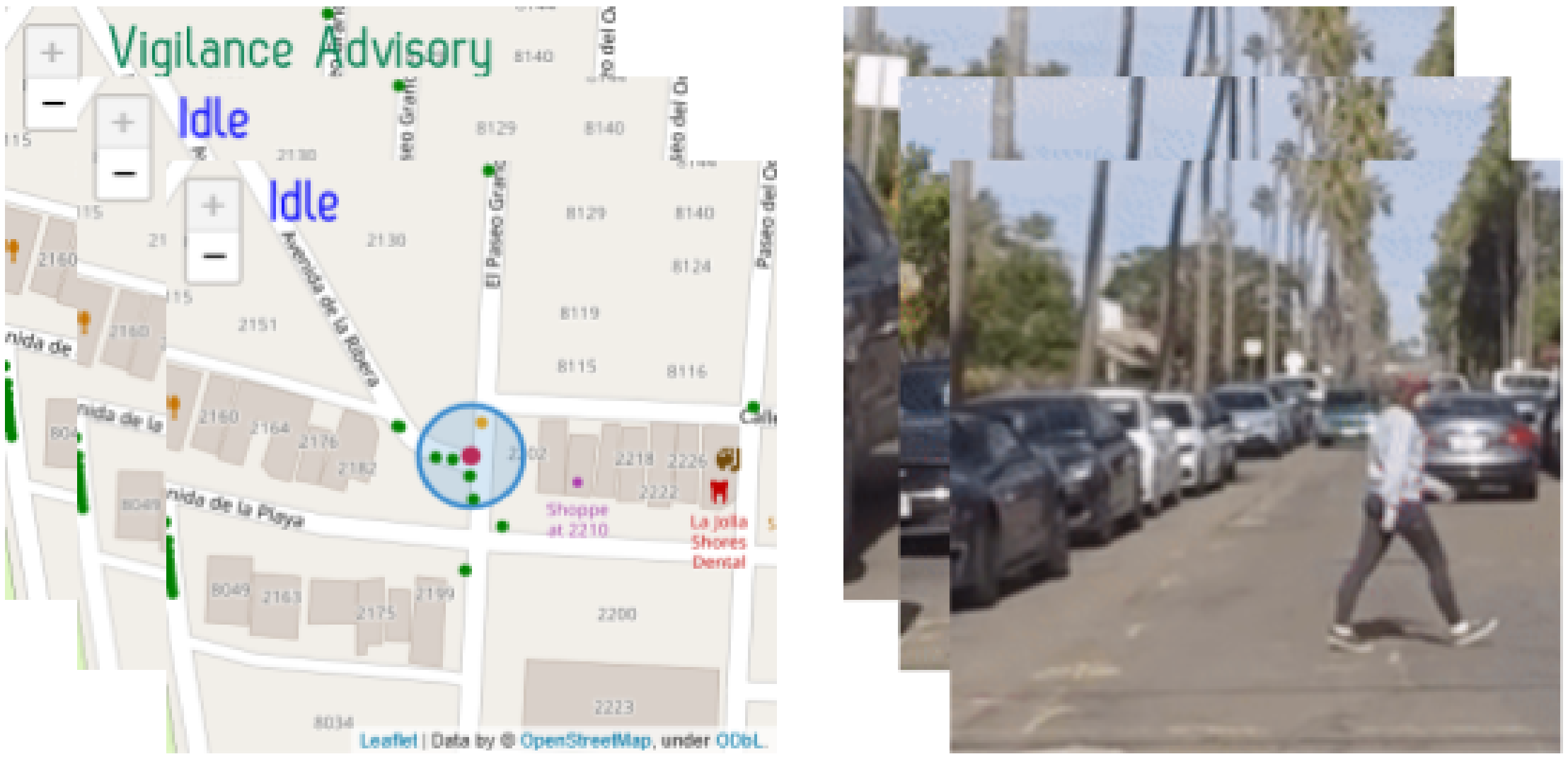

***Figure 4. CHAMP turns off vigilance advisory mode after passing all of the pedestrian training points.***

**CONCLUSIONS**

**Impact on Pedestrian Safety**

Previous research reported in [20] shows that Pedestrian Crash Avoidance and Mitigation (PCAM) systems can reduce crashes where the pedestrian crosses in front of the car or is along the side of the road by 77.6%. CHAMP's effectiveness in dark lighting conditions and when pedestrians are occluded has the potential to prevent many pedestrian incidents. Of the total 61,285 pedestrian incidents in 2020 reported in [1], 12% occurred due to visibility and 18% due to improper crossing. If CHAMP is 90% effective in such conditions, the system may prevent 16,335 pedestrian incidents per year by addressing pedestrian advisory notices in such conditions. While difficult to precisely account, if we assume that 50% of the remaining pedestrian incidents at non-intersections and intersections occur while the pedestrian is occluded, and that CHAMP is 90% effective in these conditions, we can address an additional 27,980 pedestrian incidents for cases when the pedestrian is occluded, for a total of 44,315 pedestrian incidents in the US that CHAMP may mitigate if integrated into existing navigation systems, HD maps, or automotive APIs.

**Applications and Demonstrations**

A CHAMP prototype can be demonstrated at a research conference using an interactive display, in which a user can select a pedestrian scenario and values for CHAMP hyperparameters such as sampling distance, pedestrian frequency threshold, and time of day. After selecting the hyperparameters, the user can play the clip and observe the performance of CHAMP, and compare the effects of hyperparameter choices on observed precision and recall.

In continued development, there are multiple integration modalities for CHAMP; we propose applications as a map API for existing navigation apps (for example, an option within Google Maps or Waze) for drivers in manual vehicles, and integration within autonomous vehicle navigation systems. We envision autonomous cars with state of the art 3D detectors to upload pedestrian hotspot locations to a CHAMP processing and distribution server, and in these cases, the quality of detections and the sensor suite in the autonomous car would determine how much weight or trust is given to that car's input. CHAMP can then process the information, apply denoising and thresholding, and aggregate the pedestrian sightings. For privacy and security, systems like DashCam Cleaner [21] can be used to obscure the identities of pedestrians during these detections. The aggregated sightings would be integrated into online maps which can be pulled by both traditional and intelligent cars to issue notices to the driver when entering pedestrian hotspots. The two benefits this system provides are (1) online detectors of autonomous cars can be leveraged and (2) any car or driver using a navigation system can benefit from CHAMP's safety advisories. Finally, as an ancillary application from CHAMP's maps, understanding pedestrian foot traffic can help relevant decision makers decide the road infrastructure (crosswalks, signs, etc.). CHAMP can benefit individual drivers, autonomous vehicle performance, and urban planners by estimating and continuously updating zones of high pedestrian activity.